\documentclass[letterpaper,10pt,journal,twoside]{IEEEtran}

\IEEEoverridecommandlockouts                              

\usepackage{times}
\usepackage{graphicx}
\usepackage{subfigure}
\usepackage{capt-of}
\usepackage{amsmath,amssymb,amsopn,amstext,amsfonts,amsthm}
\usepackage{cancel}
\usepackage[space]{cite}
\usepackage{pdfsync}
\usepackage{balance}
\usepackage{color}
\usepackage{mathtools}
\usepackage{bm}
\usepackage{diagbox}
\usepackage{float}
\usepackage{epstopdf}
\usepackage{pifont}
\usepackage{fixltx2e}
\usepackage{multirow}
\usepackage{url}
\usepackage{verbatim}
\usepackage[linkcolor=black,citecolor=black,urlcolor=black,colorlinks=true]{hyperref}
\usepackage{soul,xcolor}
\usepackage[linesnumbered,ruled,vlined]{algorithm2e}
\usepackage{lipsum}
\usepackage{booktabs}
\usepackage{subfiles}
\usepackage{array,booktabs}
\usepackage{mwe}
\usepackage{tablefootnote}
\usepackage{stackengine}
\newcolumntype{M}[1]{>{\centering\arraybackslash}m{#1}}

\graphicspath{{./img/}}
\DeclareGraphicsExtensions{.png,.jpg,.eps,.pdf,.jpeg}

\newcommand{\vect}{\mathbf}

\newcommand{\rev}[1]{\textcolor{black}{#1}}

\IEEEaftertitletext{\vspace{-1\baselineskip}}
\title{Active Contact Engagement for Aerial Navigation in Unknown Environments with Glass}
\author{Xinyi Chen, Yichen Zhang, Hetai Zou, Junzhe Wang and Shaojie Shen%
\thanks{Manuscript received: January 24, 2025; Revised April 2, 2025; Accepted April 28, 2025.
This paper was recommended for publication by Editor Pascal Vasseur upon evaluation of the Associate Editor and Reviewers' comments.
This work was supported by HKUST Postgraduate Studentship, Research Grants Council Research Matching Grant Scheme Project RMGS20EG20, and HKUST-DJI Joint Innovation Lab. \textit{(Xinyi Chen, Yichen Zhang and Hetai Zou contributed equally to this work.) (Corresponding author: Xinyi Chen.)}} 
\thanks{All authors are with the Department of Electronic and Computer Engineering, Hong Kong University of Science and Technology, Hong Kong, China (e-mail: $\{$xchencq,yzhangec,hzouah,jwanggj$\}$@connect.ust.hk, eeshaojie@ust.hk).}
\thanks{Digital Object Identifier (DOI): see top of this page.}
}

\IEEEaftertitletext{\vspace{-1\baselineskip}}

\begin{document}
\maketitle

\begin{abstract}
Autonomous aerial robots are increasingly being deployed in real-world scenarios, where transparent glass obstacles present significant challenges to reliable navigation. 
Researchers have investigated the use of non-contact sensors and passive contact-resilient aerial vehicle designs to detect glass surfaces, which are often limited in terms of robustness and efficiency.
In this work, we propose a novel approach for robust autonomous aerial navigation in unknown environments with transparent glass obstacles, combining the strengths of both sensor-based and contact-based glass detection. 
The proposed system begins with the incremental detection and information maintenance about potential glass surfaces using visual sensor measurements.
The vehicle then actively engages in touch actions with the visually detected potential glass surfaces using a pair of lightweight contact-sensing modules to confirm or invalidate their presence.
Following this, the volumetric map is efficiently updated with the glass surface information and safe trajectories are replanned on the fly to circumvent the glass obstacles.
We validate the proposed system through real-world experiments in various scenarios, demonstrating its effectiveness in enabling efficient and robust autonomous aerial navigation in complex real-world environments with glass obstacles.
\end{abstract}


\section{Introduction}
\label{sec:intro}
\IEEEPARstart{A}{utonomous} aerial robots are increasingly expanding their roles beyond laboratory settings, undertaking real-world applications such as autonomous exploration \cite{zhang2024falcon}, industrial inspection \cite{nikolic2013uav}, and search and rescue operations \cite{michael2014collaborative}. 
A practical scene often features glass walls and windows, but exteroceptive sensors like cameras and LiDARs can suffer from them. 
These transparent surfaces can cause unreliable sensor measurements, posing significant challenges to autonomous robot navigation in real-world scenarios.

Researchers have explored various methods to address the challenge of detecting transparent glass obstacles in the environment.
On the one hand, some studies have focused on detecting glass using a variety of non-contact sensors, including RGB cameras \cite{mei2020don}, LiDARs \cite{zhou2024lidar}, etc.
Leveraging advancements in machine learning, these methods can achieve high accuracy in detecting transparent obstacles without physical contact.
However, their performances are often influenced by factors such as lighting conditions \cite{qi2024glass} and angular uncertainty \cite{becker2012flight}.
While achieving perfect accuracy in all circumstances remains challenging, a single detection failure can block a practically viable path in the case of false positive or even cause a crash for false negative.
On the other hand, contact-resilient aerial vehicles have been designed to withstand collisions with unexpected obstacles \cite{briod2014collision, de2022collision}.
While these mechanical designs help prevent crashes, they often come with increased weight and complexity, which may compromise flight duration and narrow space maneuverability.
Although contact information has been interpreted as discrete points and utilized for mapping, numerous collisions are often required to effectively model the transparent surfaces \cite{mulgaonkar2020tiercel}, which is inefficient and impractical.
Furthermore, during autonomous navigations, these methods primarily rely on passive collision response \cite{liu2021toward}, limiting the utility of contact information as a source of sensing about the environment.
These concerns hinder the widespread adoption of autonomous aerial navigation in practical scenarios.

\begin{figure}[t]
	\centering
  \includegraphics[width=1.0\columnwidth]{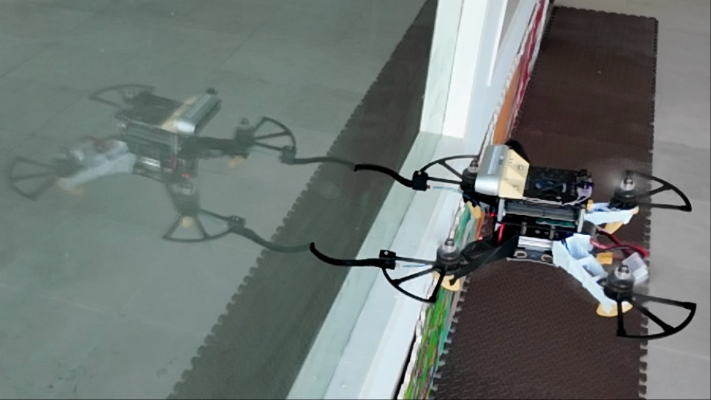}  
  \vspace{-0.6cm}
  \caption{\label{fig:touch} A real-world experiment photo captured during a touch action, when the aerial vehicle actively contacts with a visually detected glass surface to confirm its presence using a pair of front-mounted contact-sensing modules.}
  \vspace{-0.8cm}
\end{figure}

Imagine standing in front of a pristine glass door: you may first notice and suspect its presence, and then instinctively reach out to confirm it through touch.
Motivated by this natural behavior, we propose a novel approach for robust autonomous aerial navigation in complex real-world environments with transparent glass obstacles.
The proposed system combines the strengths of both sensor-based and contact-based glass detection methods, where active contact engagement provides reliable complementary information to a sensitive visual glass detector by offering true positive confirmation and false positive invalidation.
When receiving new camera images, potential glass surfaces are detected from the color image, and theirs 3D point clouds are also computed.
The information of each potential glass surface is incrementally updated during the navigation.
We integrate a lightweight contact-sensing module employing flex sensors, which provides touch sensing confirmation of the detected glass surfaces without hard collisions that risk unnecessary damage to both the agent and the glass.
For robust autonomous navigation, the proposed system is able to generate navigation trajectories that actively engage touch actions (Fig.~\ref{fig:touch}) with the visually detected glass surfaces consciously.
If a touch event is detected, the presence of the glass surface is confirmed and precisely localized using its information and the vehicle pose at the moment of contact.
The volumetric mapping is also updated on the fly, requiring only a single touch per glass surface.
Otherwise, the potential glass surface is invalidated.
We incorporate the proposed system with point-to-point aerial navigation tasks in unknown environments, aiming for more robust autonomous navigations in real-world scenes.

We evaluate the effectiveness of the incremental glass surface detection module in a glass-rich corridor, and the advantage of active contact planning compared with baselines in simulation benchmarks.
We also conduct a series of autonomous point-to-point navigation experiments in three distinct real-world scenarios to validate the proposed system.
The experimental results demonstrate the system's capability of safe and robust navigation in real-world environments with transparent glass obstacles.
In summary, the contributions are as follows:
\begin{enumerate}
  \item An incremental visual transparent glass surface detection method, that extracts glass regions from camera images and maintains essential information about potential glass surfaces in the environment.
  \item A lightweight contact-sensing module that provides touch confirmation of potential glass surfaces and invalidate false detections for non-glass surfaces, minimizing the risks associated with hard collisions while ensuring navigation efficiency.
  \item An autonomous aerial navigation system that actively engages contacts with visually detected glass surfaces for safe and robust navigation in real-world scenarios. To the best of our knowledge, this is the first work to address the challenge posed by transparent glass surfaces in aerial navigation through an active approach.
\end{enumerate}

\bstctlcite{bstctl:etal, bstctl:nodash, bstctl:simpurl}

\section{Related Work}
\label{sec:related}

\subsection{Glass Detection with Non-Contact Sensors}
Extensive research has been conducted on detecting transparent glass surfaces using a wide range of non-contact sensors.
Color cameras are among the most commonly used sensors for this purpose.
Studies in computer vision typically focus on glass identification and segmentation in a single color image, leveraging cues such as contextual information \cite{mei2020don, xie2020segmenting}, glass blurriness \cite{qi2024glass}, and boundary information \cite{he2021enhanced}.
Ultrasonic sensors offer an alternative approach for detecting glass obstacles, but they are often limited by angular uncertainty \cite{becker2012flight} and require fusion with RGBD cameras \cite{huang2018glass} or laser scan data \cite{wei2018multi} for reliable detection.
Laser rangefinders utilize the specular reflection of laser beams from the glass surface, achieving an accuracy rate of $95\%$ \cite{wang2017detecting}.
Photoelectric sensors are widely used in industrial settings to inspect transparent objects, but their limited field of view and fixed installation requirements make them unsuitable for mobile robots \cite{juds1988photoelectric}.
Polarized light cameras have also been studied for identifying transparent objects \cite{mei2022glass} and measuring depth \cite{ikemura2024robust}.
Nevertheless, they are highly susceptible to ambient lights in real-world environments and remain expensive due to limited availability.
Given the inherent limitations of non-contact sensors in achieving perfect accuracy, our work adopts a hybrid approach by combining non-contact sensor detections with physical contact to enhance the reliability of glass detection in practical applications.

\subsection{Contact-Resilient Aerial Vehicle Design}
Numerous studies have focused on developing aerial robots that can withstand collisions.
For instance, an early work \cite{briod2014collision} designs a protective frame with a gimbal system that allows it to rotate independently from the inner body.
The Tiercel \cite{mulgaonkar2020tiercel} consists of a carbon fiber frame and collision bumper to sustain collision impacts.
The ARQ \cite{liu2021toward} introduces a compliant arm design with Hall effect sensors for collision detection, which is later evolved to the s-ARQ \cite{liu2023contact} with a single compliant arm.
The RMF-Owl \cite{de2021resilient} employs a carbon-form sandwich material for main rigid component fabrication and integrates passive flaps to count collisions and adjust flight aggressiveness.
A survey \cite{de2022collision} reviews various other designs for contact-resilient aerial vehicles.
However, most of these contact-resilient aerial vehicle mechanical designs are overly complicated, heavy and cumbersome.
Although the origami-inspired design Rotorigami \cite{sareh2018rotorigami} provides a lightweight protection system, it only supports passive collision sustenance since no sensor is equipped for collision detection and characterization.
Different from these approaches, our system consciously engages gentle contact with the environment using a lightweight yet effective contact-sensing module, which is easy to integrate with most quadrotors without much modifications to the structure.

\subsection{Autonomous Navigation with Contacts and Active Methods}
With the development of contact-resilient mobile robots, researchers have investigated various roles of contacts in autonomous vehicles with different emphases. 
Some works leverage collisions to generate pseudo-measurements for localization to estimate robot states and reduce odometry uncertainty \cite{lew2019contact, li2024active}. 
Others focus on collision recovery control, enabling robots to recover from unexpected collisions and return to a normal operational state \cite{zha2020collision, lu2021deformation}.
Additionally, some studies aim to improve mapping by detecting passive collisions with unseen obstacles and estimating their positions in the map \cite{mulgaonkar2017robust, tomic2017external}.
However, only a few studies have considered utilizing contact information for autonomous navigation in unknown environments with transparent glass obstacles.
Mulgaonkar \textit{et al.} \cite{mulgaonkar2020tiercel} employ multiple passive collisions and their corresponding position to estimate the dimensions of a rectangular room with glass walls, which is inefficient and impractical for real-world applications.

Active methods have been studied for a wide range of autonomous robotics tasks since the pioneering work on active perception \cite{bajcsy1988active} in the late 1980s.
For instance, active localization drives the robot toward informative locations to improve its state estimation \cite{brunacci2024infrastructure, chen2024apace}.
Active SLAM focuses on building the most accurate and complete map of the environment \cite{cadena2016past, placed2023survey}.
Active tracking enables a robot to dynamically adjust its motion to maintain an optimal view of a target \cite{dionigi2024d, han2021fasttracker}.
To the best of the authors' knowledge, this work is the first to introduce active contact engagement for autonomous aerial navigation in glass-rich unknown environments.

\bstctlcite{bstctl:etal, bstctl:nodash, bstctl:simpurl}

\section{Incremental Glass Surface Detection}
\label{sec:mapping}
While the glass detection problem has been extensively studied in the computer vision community \cite{mei2020don, xie2020segmenting, qi2024glass, he2021enhanced}, these methods typically concentrate on a single image frame.
For the purpose of autonomous navigation, we integrate a simple yet effective visual glass detection module (Sec.~\ref{subsec:glass_detection}) that can be deployed on edge devices for real-time processing.
Furthermore, we develop an incremental approach (Sec.~\ref{subsec:glass_update}) that fuses potential glass surface detections from consecutive frames, maintaining the information of glass surfaces in the environment in a data structure detailed in Table \ref{tab:glass_data}.

\begin{table}[t]
  \centering
  \caption{Information Stored in a Glass Surface Data Structure $G_i$}
  \vspace{-0.2cm}
  \begin{tabular}{cc}
    \toprule\toprule
    \textbf{Notation} & \textbf{Description} \\
    \midrule
    $\vect{c}_i$ & Centroid of the glass surface\\
    $\vect{n}_i$ & Normal of the glass surface\\
    $P_i$ & Polygon vertices of the glass surface boundary\\
    $C_i$ & Point cloud of the glass surface\\
    \toprule\toprule
  \end{tabular}
  \label{tab:glass_data}
  \vspace{-0.8cm}
\end{table}

\subsection{Visual Glass Detection}
\label{subsec:glass_detection}

On transparent glass obstacles, exteroceptive sensors like cameras and LiDARs can produce noisy or invalid measurements, due to a lack of opaque and detectable features.
For autonomous navigation, we first need to efficiently detect potential glass surfaces from individual frames of onboard RGBD camera images.

\begin{figure}[t]
  \centering
  \subfigtopskip=0pt
  \subfigbottomskip=0pt
  \subfigcapskip=-3pt
  \subfigure[]{\includegraphics[width=0.48\columnwidth]{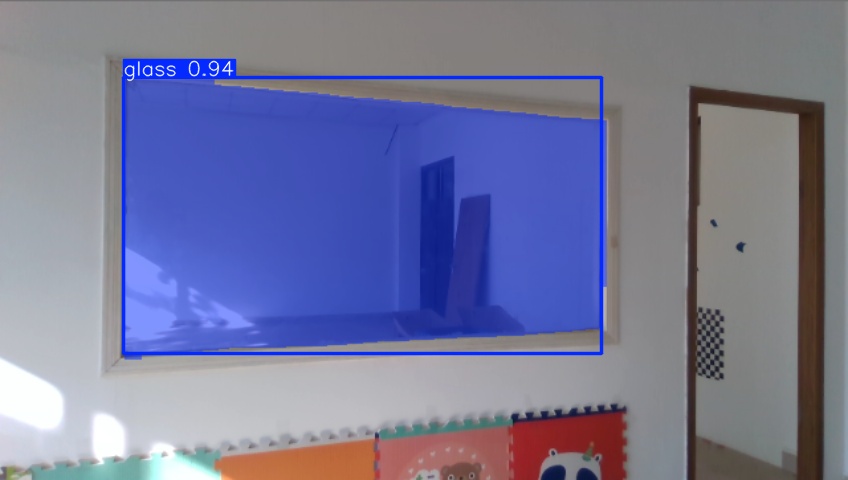}} 
  \subfigure[]{\includegraphics[width=0.48\columnwidth]{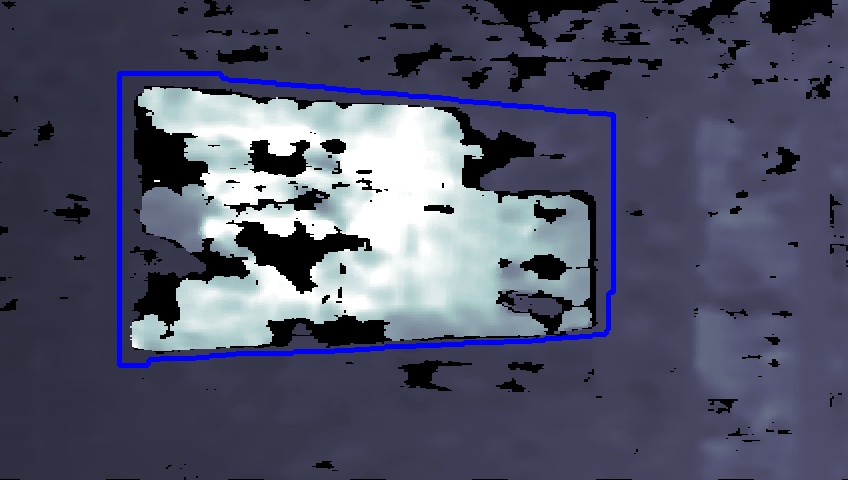}}\\
  \subfigure[]{\includegraphics[width=0.48\columnwidth]{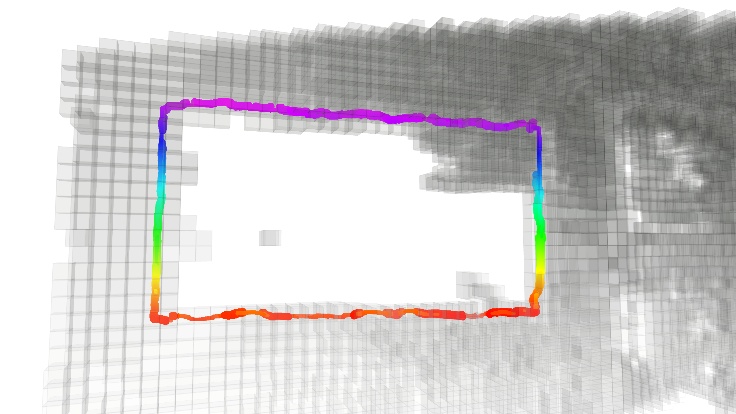}}     
  \subfigure[]{\includegraphics[width=0.48\columnwidth]{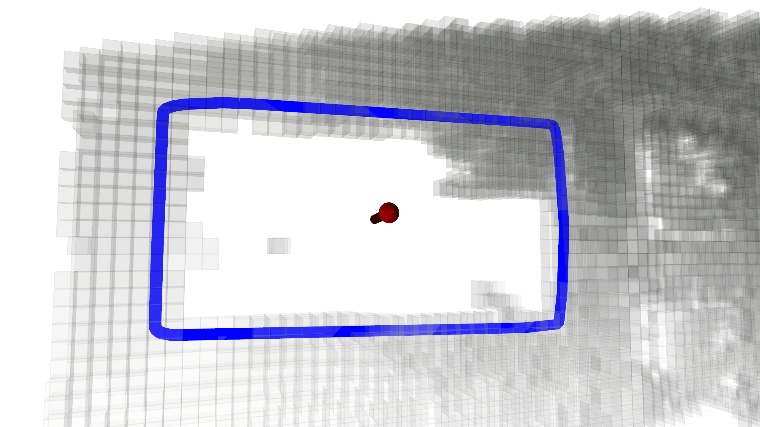}}
  \caption{\label{fig:window_detection} An illustration of the proposed visual glass surface detection module. (a) Glass segmentation from the color image. (b) Boundary points in the depth image. (c) The 3D point set of the glass surface boundary in the world frame. (d) The polygon of the glass surface on the fitted plane. In (c)(d), the grey voxels are occupancy grids constructed from the depth image.}
  \vspace{-0.8cm}
\end{figure}

The proposed visual glass surface detection module is detailed in Line~\ref{alg_line:detection_start}-\ref{alg_line:detection_end} of Alg.~\ref{alg:window_detection}.
We observe that though glass surfaces are transparent, they often have visible edges and frames that can produce reliable measurements.
Hence we attempt to utilize these boundaries to detect and extract glass surfaces from RGBD images.
Firstly, the regions of glass segmentations $\mathcal{S}$ are extracted from the color image $\mathcal{I}$ with an object detection and instance segmentation framework YOLOv8 \cite{yolov8ultralytics}. 
A segmentation with a confidence score exceeding a threshold $\tau_s$ is valid and considered as a potential glass surface (Fig.~\ref{fig:window_detection}(a)).
The boundary of each segmentation is dilated to obtain a set of boundary points $H_i$ with stable depth measurements.
We query the pixels of the boundary points in $H_i$ in the synchronized and aligned depth image $\mathcal{D}$ (Fig.~\ref{fig:window_detection}(b)) and project them to the 3D world coordinates to obtain point set $Q_i$ (Fig.~\ref{fig:window_detection}(c)).
Due to the noisy depth measurements, $Q_i$ is further fitted into a plane $\pi_i$ using the least square method RANSAC \cite{Fischler1981RandomSC}.
The inlier points are projected to the fitted plane $\pi_i$ to obtain $\bar{Q}_i$. 
The convex hull of $\bar{Q}_i$ is then computed, yielding the glass surface boundary as a polygon defined by its vertices $P_i$ (Fig.~\ref{fig:window_detection}(d)).
The centroid $\vect{c}_i$ of the glass surface is computed as the average of the points in $P_i$, and the normal $\vect{n}_i$ of the glass surface is estimated as the normal of the fitted plane $\pi_i$.
Lastly, a 3D point cloud $C_i$ is sampled within the polygon formed by $P_i$ to represent the glass surface obstacle for volumetric mapping refinement in later stages.
At this point, we have obtained the essential information of each potential glass surface $G_i$ contained in the current single image frame.

\subsection{Incremental Glass Surface Update Across Frames}
\label{subsec:glass_update}

As the glass surfaces may not be fully observed in a single frame, fusing the detection results across different frames becomes crucial for maintaining information of the glass surfaces in the environment.
Whenever new sensor measurements are received during the navigation, results $G_i$ from the visual glass detection in Sec.~\ref{subsec:glass_detection} should be aggregated with the existing glass surface list $\mathcal{G}$ incrementally.
The overall incremental update process is described in Line~\ref{alg_line:update_start}-\ref{alg_line:update_end} of Alg.~\ref{alg:window_detection}.

To fuse the detection results across different frames, data association is required to recognize whether the incoming glass surface $G_i$ is an observation of one of the existing surfaces in $\mathcal{G}$ or a new glass surface.
For each newly detected glass surface $G_i$ and each existing $G_j \in \mathcal{G}$, we first compute the angle between their normal directions $\theta(\vect{n}_i, \vect{n}_j)$ and the Euclidean distance between their centroids $d(\vect{c}_i, \vect{c}_j)$.
An angle exceeding a threshold $\tau_n$ or a distance exceeding $\tau_c$ indicates that they are different surfaces.
For surfaces with similar normal directions and centroids, we further compute their Intersection over Union (IoU) score $IoU(i,j)$.
We project points in $P_i$ to the plane holding $G_j$ to obtain $^jP_i$.
Then we compute the 2D IoU score $IoU(i,j)$ as
$IoU(i,j) = \text{area}(\mathcal{P}_i \cap {^i\mathcal{P}_j}) / \text{area}(\mathcal{P}_i \cup {^i\mathcal{P}_j}),$
where the polygons $^j\mathcal{P}_i$ and $\mathcal{P}_j$ are defined by vertices $^jP_i$ and $P_j$.
For each incoming glass surface $G_i$, the highest IoU score $\hat{IoU}(i)$ and its index $k$ can be computed by
\begin{equation}
  \hat{IoU}(i) = \mathop{\max}_{j} IoU(i,j), \quad k = \arg \mathop{\max}_{j} IoU(i,j).
\end{equation}
In the case that the highest IoU score $\hat{IoU}(i)$ is lower than a threshold $\tau_{iou}$, we treat $G_i$ as a new potential glass surface and add it to the list $\mathcal{G}$.
Otherwise, $G_i$ and $G_k$ are considered to be observations for the same glass surface. 
Hence, we update the information for the existing glass surface $G_k$ using the new detection $G_i$.
A new fitted plane $\pi_k^\prime$ for the glass surface $G_k$ is computed using points from $P_i$ and $P_k$, following a process similar to Sec.~\ref{subsec:glass_detection}. The normal of this plane is considered the updated normal $\vect{n}_k^\prime$ of the glass surface $G_k$. 
The updated polygon vertices of $G_k$ are then computed as
\begin{equation}\label{eq:union}
  P_k^\prime = {^{k^\prime}P_i} \cup {^{k^\prime}P_k}, 
\end{equation}
where $^{k^\prime} P_i$ and $^{k^\prime} P_k$ are the projection of $P_i$ and $P_k$ onto the plane defined by the updated normal $\vect{n}_k^\prime$ and centroid $\vect{c}_k^\prime$. 
The union function in (\ref{eq:union}) calculates the spatial set theoretic union that combines two polygons into one rather than a simple concatenation of the vertices.
Finally, the updated centroid $\vect{c}_k^\prime$ is computed as the average of the points in $P_k^\prime$.

\begin{algorithm}[t]
  \label{alg:window_detection}
  \caption{Incremental Glass Surface Detection}\label{alg:two}
  \KwIn{RGB image $\mathcal{I}$, depth image $\mathcal{D}$}
  \KwOut{Updated glass surfaces list $\mathcal{G}$}
  \BlankLine
  
  \SetKwFunction{Segmentation}{Segmentation} 
  \SetKwFunction{ConvexHull}{ConvexHull}
  \SetKwFunction{MaskBoundary}{MaskBoundary}
  \SetKwFunction{ProjectToWorld}{ProjectToWorld}
  \SetKwFunction{FitPlaneRANSAC}{FitPlaneRANSAC}
  \SetKwFunction{ProjectToPlane}{ProjectToPlane}
  \SetKwFunction{ComputeMaxIoU}{ComputeMaxIoU}
  \SetKwFunction{update}{update}
  
  $\mathcal{G}_\text{new} \gets \emptyset$  \label{alg_line:detection_start}\\
  $\mathcal{S}$ $\gets$ \Segmentation{$\mathcal{I}$}\\
  
  \ForEach{\textnormal{glass sementation mask} $s_i \in \mathcal{S}$}{
    $H_i \gets$ \MaskBoundary{$s_i$} \\
    \ForEach{\textnormal{boundary point} $\vect{h}_j$ $\in$ $H_i$}{
        $d_j$ $\gets \mathcal{D}(\vect{h}_j.x, \vect{h}_j.y)$ \\
        $\vect{q}_j$ $\gets$ \ProjectToWorld{$\vect{h}_j, d_j$}\\
        $Q_i$.add($\vect{q}_j$)
    }
    $\pi_i$ $\gets$ \FitPlaneRANSAC{$Q_i$}\\
    $\bar{Q}_i$ $\gets$ \ProjectToPlane($Q_i$, $\pi_i$)\\
    $P_i$ $\gets$ \ConvexHull{$\bar{Q}_i$} \\
    $\vect{c}_i, \vect{n}_i$ $\gets$ $P_i$.centroid, $\pi_i$.normal\\
    $C_i$ $\gets$ \textnormal{samplePointCloud}($P_i$)\\

    $G_i$ $\gets$ \textit{new} GlassSurface($\vect{c}_i, \vect{n}_i, P_i, C_i$)\\
    $\mathcal{G}_\text{new}$.add($G_i$) \label{alg_line:detection_end}
  } 
  
  \ForEach{$G_i \in \mathcal{G}_\text{new}$ \textnormal{and} $G_j \in \mathcal{G}$}{ \label{alg_line:update_start}
      $^jP_i$ $\gets$ \ProjectToPlane($P_i$, $G_j$.plane) \\
      $\hat{IoU}(i)$, $k$ $\gets$ \ComputeMaxIoU($P_j$, $^jP_i$) \\
      \If {$\hat{IoU}(i) < \tau_{iou}$} {
        $\mathcal{G}$.add($G_i$) \\
      }
      \Else {
        $G_k$.\update($G_i$) \\
      }
      
  }\label{alg_line:update_end}
\end{algorithm}

\section{Active Contact Engagement in Autonomous Aerial Navigation}
\label{sec:method}
Building on the extracted glass surface information from camera images, the proposed system can next confirm their presence by purposefully contacting the detected glass using a lightweight contact-sensing module (Sec.~\ref{subsec:flex}). 
Utilizing touch actions (Sec.~\ref{subsec:touch_action}), we integrate active contact engagement into a fundamental autonomous point-to-point aerial navigation task (Sec.~\ref{subsec:navigation}). 

\subsection{Contact-Sensing Module}
\label{subsec:flex}
\begin{figure}[t]
	\centering
  \subfigure[]{\includegraphics[height=0.38\columnwidth]{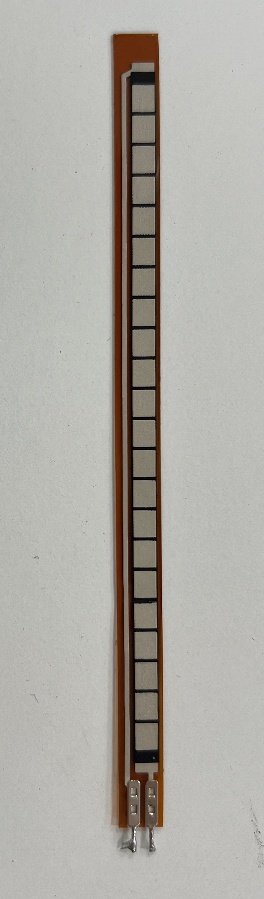}}
  \subfigure[]{\includegraphics[height=0.38\columnwidth]{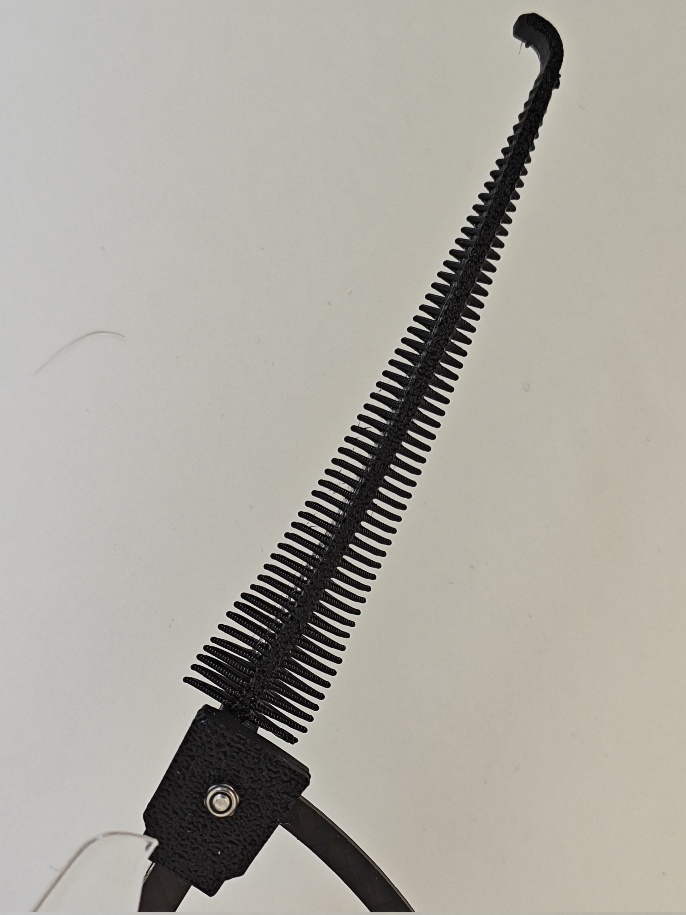}}
  \subfigure[]{\includegraphics[height=0.38\columnwidth]{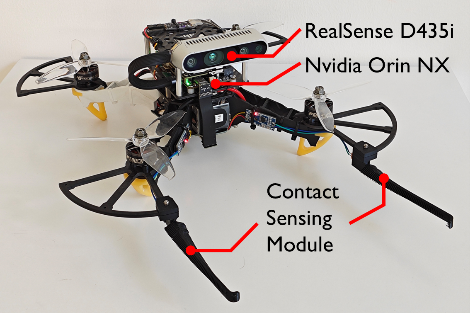}}
  \vspace{-0.4cm}
  \caption{\label{fig:flex} The figure illustrates (a) the flex sensor used in our design, (b) the proposed contact-sensing module, (c) the aerial vehicle for flight experiments.}
  \vspace{-0.8cm}
\end{figure}

We design a lightweight contact-sensing module mounted in the front of the aerial vehicle to detect physical contact with environmental obstacles, as shown in Fig.~\ref{fig:flex}.
The module consists of a $4.5$-inch flex sensor that measures deflection and is encased in a 3D-printed TPU shell.
Upon contact with obstacles, the contact-sensing module bends physically, causing the resistance of the flex sensor to change.
This resistance is converted into a voltage signal, digitized by a calibrated analog-to-digital converter, and transmitted to the high-level onboard computer.
The soft and deformable TPU shell serves two purposes: it protects the flex sensor from fractures due to excessive bending and acts as a suspension system to mitigate the effects of aerial vehicle body vibrations and air turbulence on the flex sensor.
This design ensures the sensor remains durable and functional during physical contact with obstacles.
Weighting only $3.3$ g each, the lightweight contact-sensing module is portable and easy to install on various aerial platforms.
Unlike hard collisions, the proposed contact-sensing module introduces negligible impact on the aerial vehicle during interactions, reducing the risk of damage to the vehicle.
Additionally, a previous study \cite{mulgaonkar2020tiercel} has revealed the correlation between collision impact and state estimation failure.
By enabling gentle touch sensing, the module enhances the survivability of onboard state estimation, further improving the robustness of autonomous navigation.

\subsection{Touch Action for Active Contact Engagement}
\label{subsec:touch_action}
Before discussing autonomous navigation with active contact engagement, we first introduce how a \textit{touch action} is performed when the aerial vehicle intentionally approaches and attempts to make contact with a target potential glass surface.
Assume the target surface has a centroid $\vect{c}$ and a unit normal vector $\vect{n}$ pointing towards the aerial vehicle.
The touch action begins at a ready pose, defined by the position $\vect{c} + \delta_s \vect{n}$, where $\delta_s$ is a predefined distance, and an orientation that aligns the contact-sensing module with the target surface.
The planner then generates a smooth trajectory from the ready pose to the ending position $\vect{c} - \delta_e \vect{n}$, maintaining a constant yaw and a conservative touch speed $v_\text{touch}$ to ensure gentle contact.
During the flight, if the contact-sensing module detects a touch event, the aerial vehicle confirms the presence of the glass surface.
The contact position can be computed using the vehicle's pose and the physical parameters of the contact-sensing module.
This contact position provides precise information about the exact location of the glass surface.
Subsequently, the volumetric map is updated using the glass surface information, where its point cloud $C_i$ is offset along the surface normal $\vect{n}_i$ to ensure the contact position lies on the surface.
The aerial vehicle then returns to the ready pose for safety and subsequent replanning.
If the aerial vehicle reaches the ending position without triggering the contact-sensing module, it records the potential glass surface as invalid and proceeds with the navigation.
A photo capturing the moment of a touch action is shown in Fig.~\ref{fig:touch}.

\subsection{Autonomous Navigation with Active Contact Engagement}
\label{subsec:navigation}
Active contact engagement enhances the safety of point-to-point navigation in unknown environments, particularly when visual sensor measurements of transparent glass surfaces are unreliable.
In such cases, the occupancy map may incompletely represent glass surfaces, increasing the risk of crashes if the aerial vehicle attempts to traverse these areas.
Equipped with the visual detection module and contact-sensing capability, the aerial vehicle obtains awareness of transparent glass obstacles in the environment during autonomous navigation.

\begin{figure}[t]
	\centering
  \includegraphics[width=0.75\columnwidth]{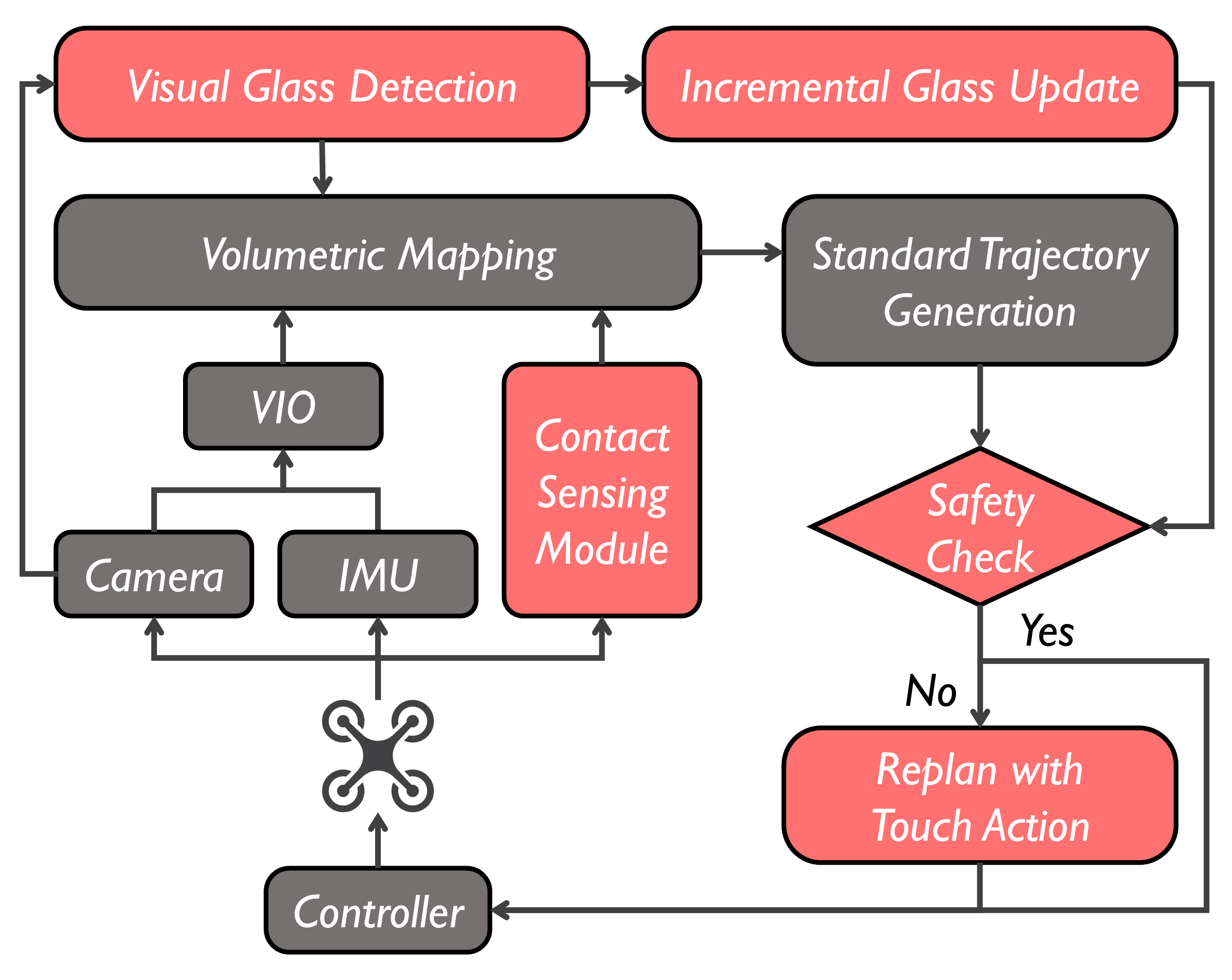}  
  \vspace{-0.4cm}
  \caption{\label{fig:system} The figure illustrates the pipeline of the proposed autonomous navigation system, with the pink blocks highlighting components related to active contact engagement.}
  \vspace{-0.8cm}
\end{figure}

\begin{figure}[t]
	\centering
  \includegraphics[width=\columnwidth]{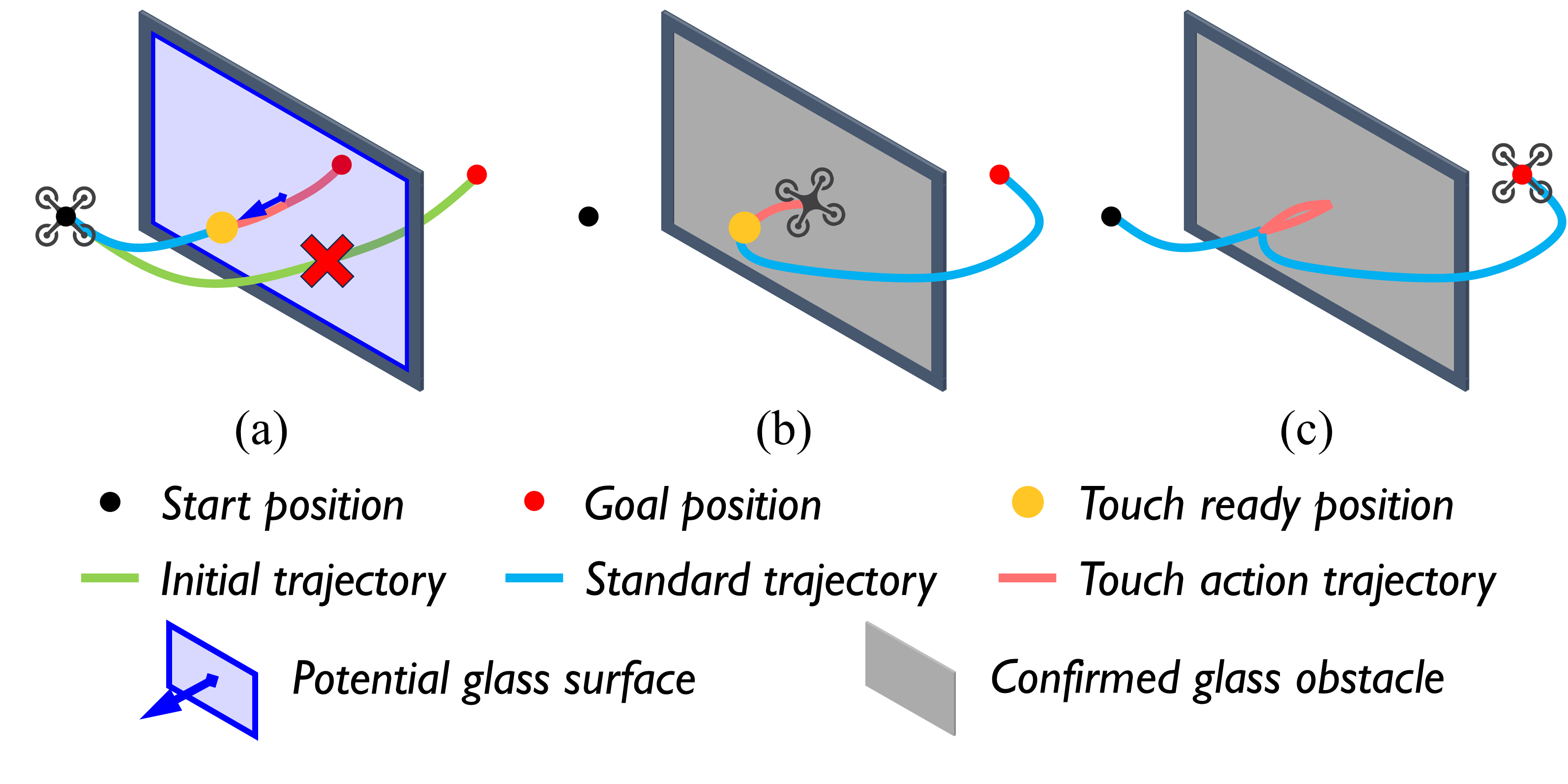}  
  \vspace{-1.0cm}
  \caption{\label{fig:p2p} An illustration of the proposed autonomous navigation system with active contact engagement in a point-to-point aerial navigation task.}
  \vspace{-0.8cm}
\end{figure}

This section describes the integration of active contact engagement into the point-to-point aerial navigation task in unknown environments.
Based on a standard trajectory planner, the diagram for the proposed autonomous navigation system with active contact engagement is shown in Fig.~\ref{fig:system}.
Given a start and a goal position, the planner first generates a standard trajectory using the volumetric map derived from conventional sensor measurements.
A safety check is then performed to identify any intersection with potential glass surfaces by sampling along the trajectory (green trajectory in Fig.~\ref{fig:p2p}(a)).
If the trajectory is safe without intersection with any potential glass surface, the aerial vehicle follows the standard trajectory to the goal position.
Otherwise, the planner generates a new trajectory to the ready pose of the first intersected surface and decelerates the vehicle to $v_\text{touch}$ to execute a touch action (blue trajectory in Fig.~\ref{fig:p2p}(a)).
During the touch action (Sec.~\ref{subsec:touch_action}), the volumetric map is updated with the contact information (Fig.~\ref{fig:p2p}(b)).
Once the touch action is completed and the glass surface is confirmed or invalidated, the planner replans another standard trajectory to the goal position using the updated volumetric map (Fig.~\ref{fig:p2p}(c)).
This process repeats iteratively until the goal position is reached.
Note that the point-to-point navigation task represents only one of many potential applications for the proposed system.
It can be extended to more comprehensive tasks by replacing the standard trajectory generation component in Fig.~\ref{fig:system} with a corresponding planner tailored to the specific task.

\section{Experiments and Results}
\label{sec:exp}

\subsection{Implementation Details}
\label{subsec:exp_impl}
For the onboard software, the object segmentation framework YOLOv8 \cite{yolov8ultralytics} utilized in Sec.~\ref{subsec:glass_detection} is custom-trained using transparent surface data aggregated from open-source datasets \cite{linExploitingSemanticRelations2022, mei2020don, linRichContextAggregation2021}.
The spatial set theoretic union in Sec.~\ref{subsec:glass_update} leverages the Boost C++ Libraries\footnote{\url{https://beta.boost.org/doc/libs/1_58_0/libs/geometry/doc/html/geometry/reference/algorithms/union_.html}}.
The standard trajectory for the point-to-point navigation task in Sec.~\ref{subsec:navigation} is generated using the planner proposed in \cite{zhou2019robust}.
Table~\ref{tab:params} summarizes the parameters setting for the proposed algorithm, which is used throughout the experiments.

In the real-world flight experiments (Sec.~\ref{subsec:exp_glass} to \ref{subsec:exp_window}), we customize the contact-sensing module based on an open-source lightweight quadrotor platform \cite{zhang2024uniquad}, as shown in Fig.~\ref{fig:flex}(c).
The quadrotor is equipped with an NVIDIA Jetson Orin NX\footnote{\url{https://www.nvidia.com/en-us/autonomous-machines/embedded-systems/jetson-orin}} 16 GB, a NxtPx4 autopilot \cite{liu2024omninxt}, a global-shutter stereo camera RealSense D435i\footnote{\url{https://www.intelrealsense.com/depth-camera-d435i}} and a pair of contact-sensing modules (Sec.~\ref{subsec:flex}), which has a take-off weight of $895$ g.
The linear and angular dynamics limitations of the quadrotor are set to $v_m = 1.0$ m/s, $a_m = 1.0$ m/s$^2$, $\dot{\xi}_m = 1.05$ rad/s, $\ddot{\xi}_m = 1.05$ rad/s$^2$.
The localization of the quadrotor is provided by a visual-inertial state estimator VINS-Fusion\cite{qin2019general}.
All the experiments are conducted fully onboard without other external infrastructure, such as ground stations or motion capture systems.

\subsection{Incremental Glass Detection in a Glass-Rich Corridor}
\label{subsec:exp_incremental}
\begin{figure}[t] 
	\centering 
  \subfigtopskip=0pt
	\subfigbottomskip=0pt 
	\subfigcapskip=-3pt
  \subfigure[]{\includegraphics[width=0.43\columnwidth]{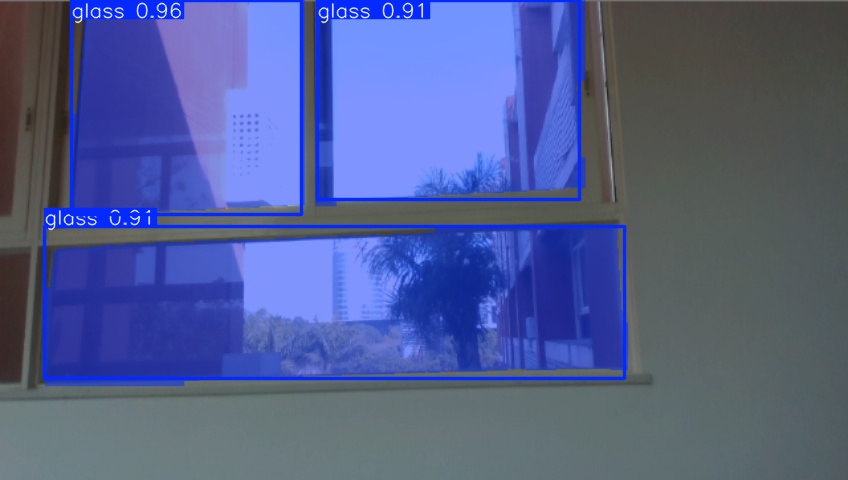}}
  \subfigure[]{\includegraphics[width=0.43\columnwidth]{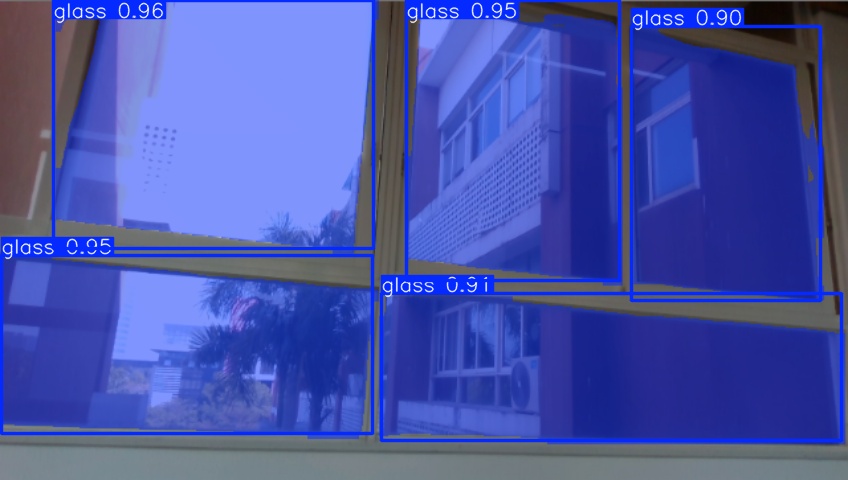}}   
  \subfigure[]{\includegraphics[width=0.43\columnwidth]{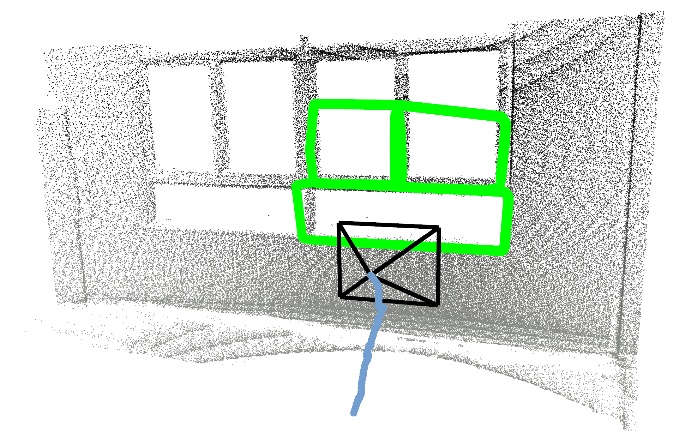}} 
  \subfigure[]{\includegraphics[width=0.43\columnwidth]{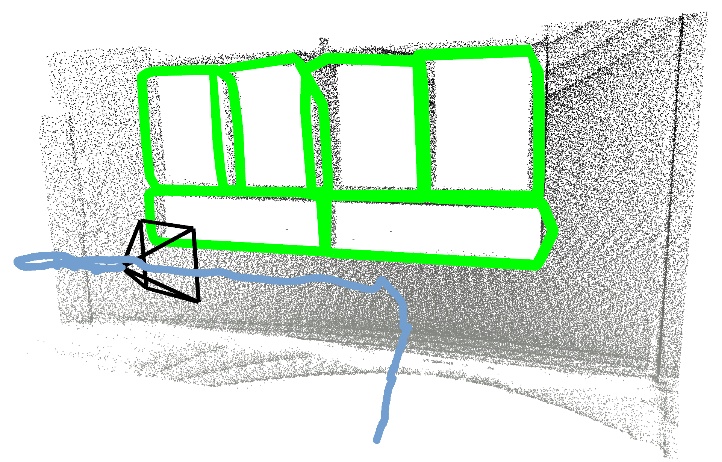}} 
  \caption{\label{fig:window_mapping} Experiment results of incremental glass surface detection. The left and right columns showing the initial and final stages of the experiment respectively. The upper row shows the first-person view alongside the visual segmentation results. The lower row displays the trajectory and results, with the grey point cloud for visualization and ground truth comparison.}
  \vspace{-1.2cm}
\end{figure}

We first conduct a unit testing on the incremental glass surface detection capability of the proposed system.
In this experiment, data is collected in a glass-rich corridor using a handheld device equipped with a RealSense D435i camera and a Livox Mid-360 LiDAR\footnote{\url{https://www.livoxtech.com/mid-360}}.
Initially, the agent can only capture a partial view of the glass windows, as shown in Fig.~\ref{fig:window_mapping}(a)(c).
With more measurements received, the visual detection module incrementally updates the detected glass surfaces in the scene, completing the glass surface detection of all glass windows, as shown in Fig.~\ref{fig:window_mapping}(b)(d).
Please note that the LiDAR is used solely for reconstructing the scene for ground truth visualization and its data is not used in the proposed system.
The experiment results demonstrate that the proposed glass detection module is capable of incrementally updating glass surface information in real time.

\begin{table}[t]
  \caption{Parameters Setting}
  \vspace{-0.2cm}
    \centering
    \begin{tabular}{cccc}
    \toprule\toprule
    \textbf{Parameter} & \textbf{Section} & \textbf{Notation} & \textbf{Value} \\
      \midrule
      Segmentation confidence & \ref{subsec:glass_detection} & $\tau_s$ & $0.75$ \\
      Normal difference threshold & \ref{subsec:glass_update} & $\tau_n$ & $0.65$ rad \\
      Centroid difference threshold & \ref{subsec:glass_update} & $\tau_c$ & $1.0$ m \\
      IoU threshold & \ref{subsec:glass_update} & $\tau_i$ & $0.1$ \\
      Touch action start distance & \ref{subsec:touch_action} & $\delta_s$ & $1.0$ m\\
      Touch action end distance & \ref{subsec:touch_action} & $\delta_e$ & $1.0$ m\\
      Touch action velocity & \ref{subsec:touch_action} & $v_\text{touch}$ & $0.2$ m/s\\
      \toprule\toprule
  \end{tabular}
  \label{tab:params}
  \vspace{-0.2cm}
\end{table}

\subsection{Simulation Benchmark for Aerial Navigation}
\label{subsec:exp_planning}
We conducted benchmark simulations to evaluate the proposed active contact navigation planner against two baseline methods:
(1) \textit{Non-contact}: A vision-only planner that treats all visually detected glass surfaces as obstacles without physical interaction, and
(2) \textit{Contact-based}: An adaptation of a state-of-the-art planner designed for glass-rich environments \cite{mulgaonkar2020tiercel} that performs passive contact reactions and marks regions near contact points as occupied space.

\begin{figure}[t] 
	\centering 
  \subfigtopskip=0pt
	\subfigbottomskip=2pt 
	\subfigcapskip=-6pt
  \subfigure[Non-contact]{\includegraphics[width=0.32\columnwidth]{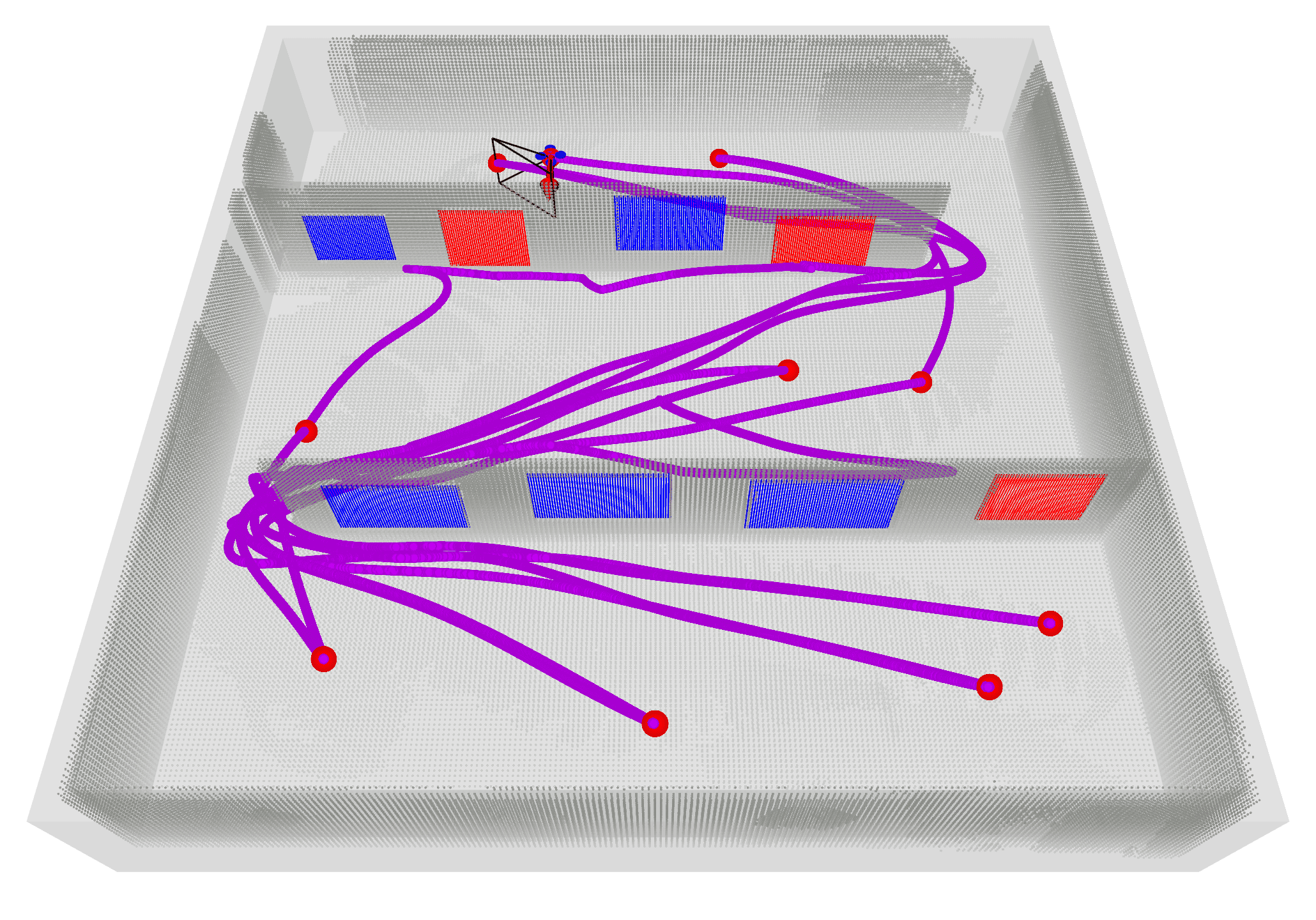}}
  \subfigure[Contact-based]{\includegraphics[width=0.32\columnwidth]{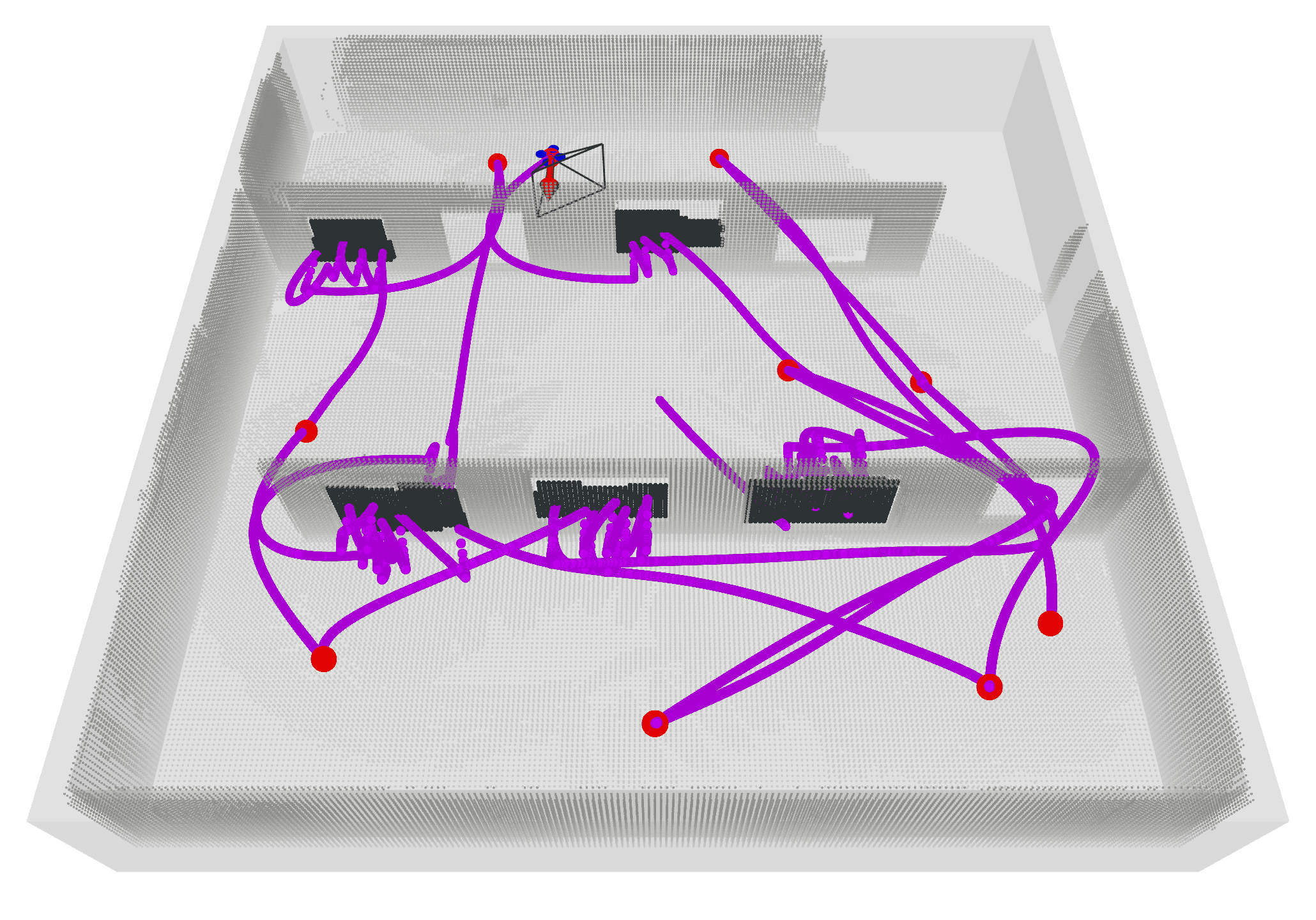}} 
  \subfigure[Proposed]{\includegraphics[width=0.32\columnwidth]{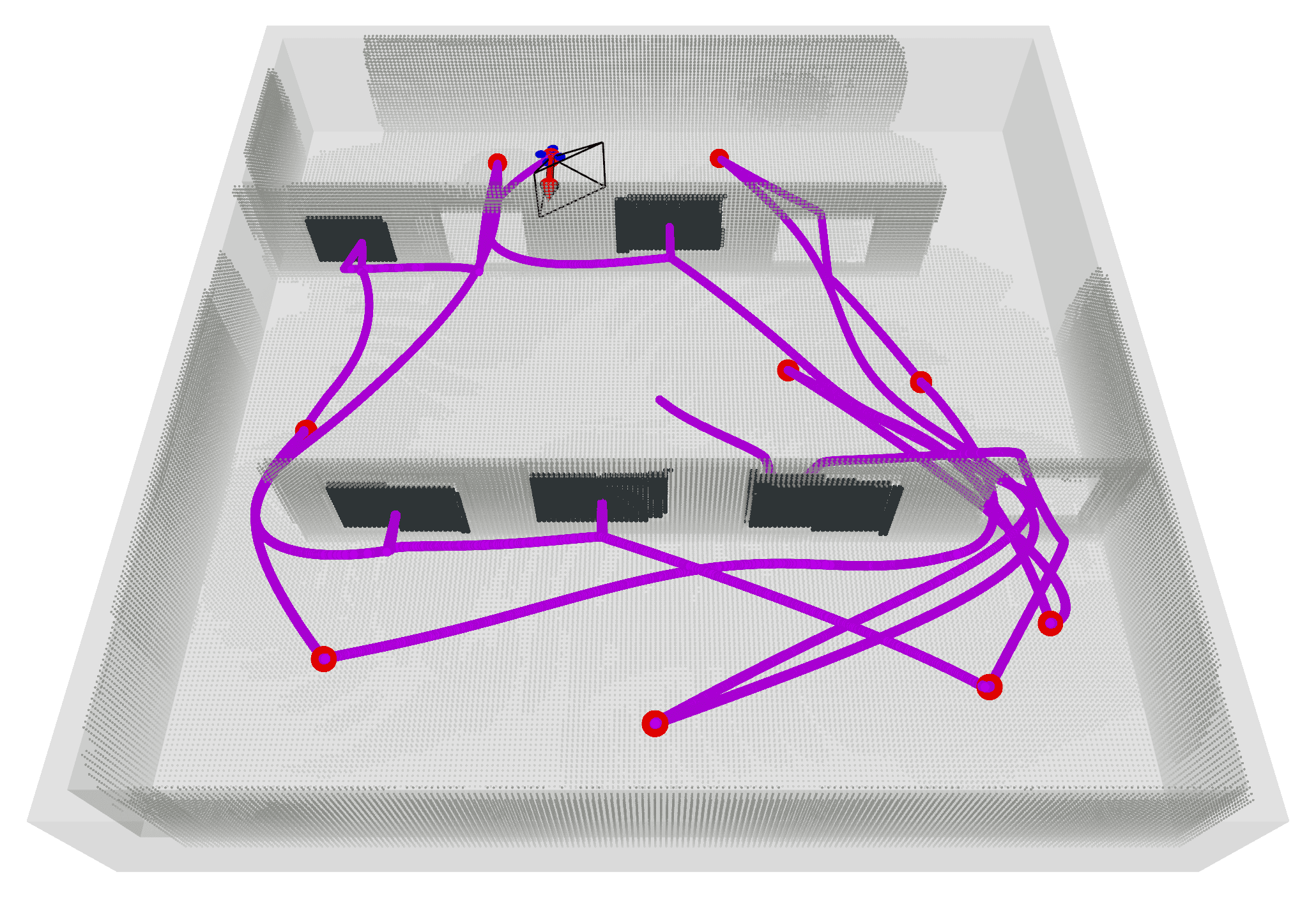}} 
  \caption{\label{fig:sim} The executed trajectories in simulation benchmark experiments. The red dots indicate the randomly sampled target positions.}
  \vspace{-1.2cm}
\end{figure}

Due to simulation constraints in rendering realistic camera measurements on transparent and reflective glass, the 2D visual glass detection and segmentation is directly provided by the simulator for all methods.
As illustrated in Fig.~\ref{fig:sim}(a), blue regions indicate correctly detected glass surfaces (true positives), and red regions show the traversable openings that are incorrectly detected as glass (false positives), which are both only for visualization purposes. The planners are required to perform real-time mapping of occupancy map and glass surfaces during navigation.
The experiment consists of $10$ point-to-point navigation tasks, with randomly sampled non-linear target positions.
Each method was tested for $5$ runs, with quantitative results summarized in Table~\ref{tab:sim} and representative trajectories shown in Fig.~\ref{fig:sim}.
The results demonstrate that the non-contact visual-only method suffered from conservative detours, while the contact-based method required numerous passive contacts to fully reconstruct the glass surfaces. 
In contrast, The proposed active contact method consciously confirmed or invalidated the detected glass surfaces in a single touch, yielding the shortest flight time and path length among the baselines.

\begin{table}[t]
  \caption{Simulation Benchmark Result Statistics}
  \vspace{-0.4cm}
  \begin{center}
    \begin{tabular}{ccccccc}
      \toprule\toprule
      \multirow{2}{*}{\textbf{Method}} & \multicolumn{2}{c}{\textbf{Duration (s)}} & \multicolumn{2}{c}{\textbf{Path Length (m)}} & \multicolumn{2}{c}{\textbf{\#Contact$^\dagger$}}\\
                                       & \textbf{Avg} & \textbf{Std} & \textbf{Avg} & \textbf{Std} & \textbf{Avg} & \textbf{Std} \\
      \midrule
      Non-contact & 282.0 & 8.4 & 288.3 & 6.1 & - & -\\
      Contact-based & 265.2 & 13.7 & 248.9 & 7.7 & 35.6 & 3.4\\
      Proposed & \textbf{223.2} & 2.7 & \textbf{159.7} & 0.5 & 5.2 & 0.4\\
      \toprule\toprule
    \end{tabular}
  \end{center}
  \vspace{-0.2cm}
  \footnotesize{$^\dagger$ The number of physical contacts with glass surfaces.}\\
  
  \label{tab:sim}
  \vspace{-0.4cm}
\end{table}

\subsection{Navigating a Space with a Glass Door}

\label{subsec:exp_glass}
\begin{figure}[t] 
	\centering 
  \subfigure[]{\includegraphics[height=0.27\columnwidth]{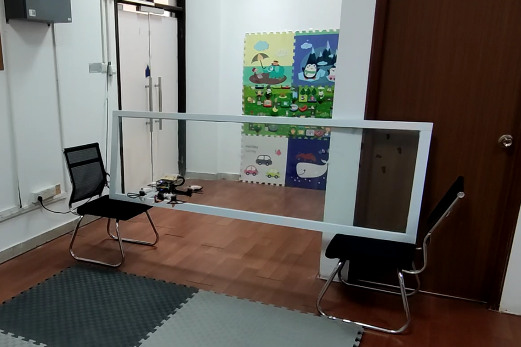}}   
  \subfigure[]{\includegraphics[height=0.27\columnwidth]{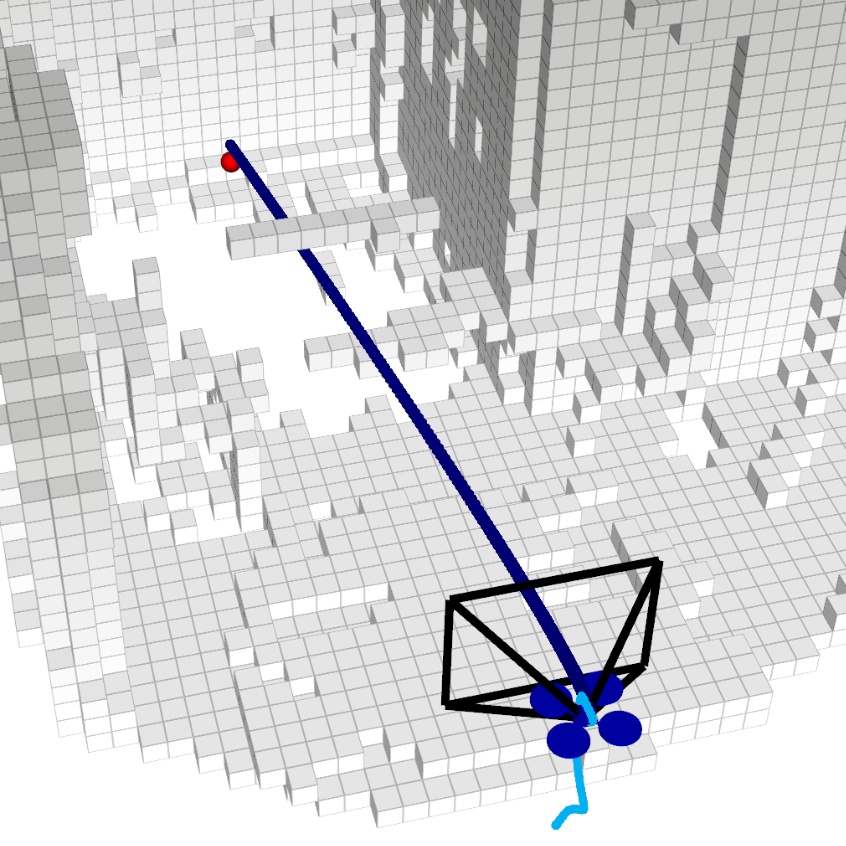}} 
  \subfigure[]{\includegraphics[height=0.27\columnwidth]{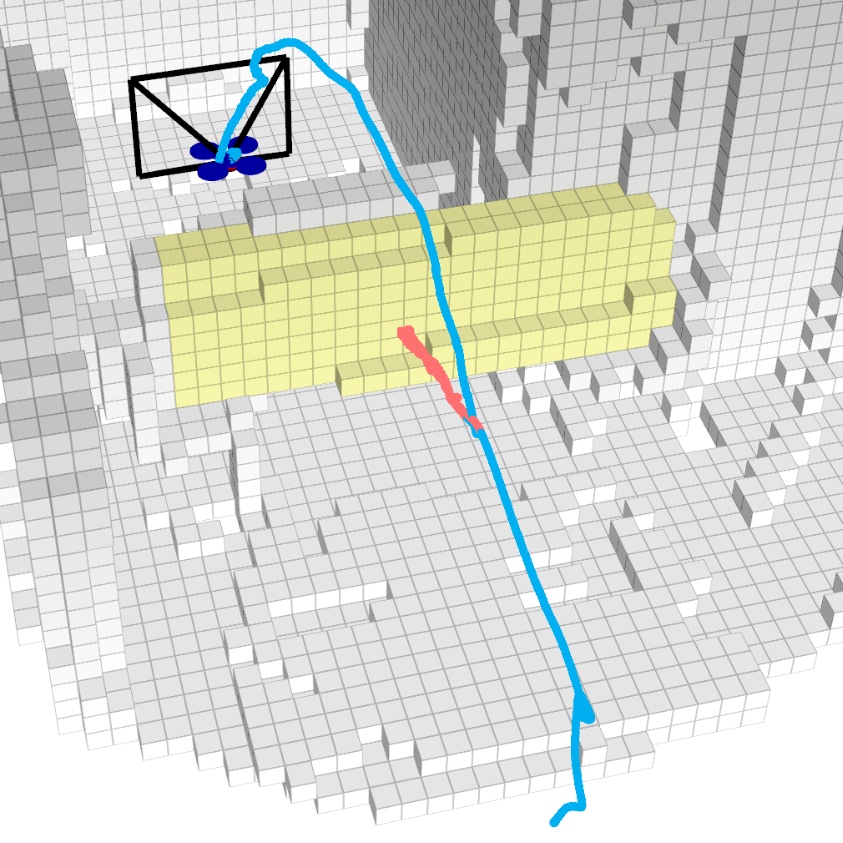}} 
  \vspace{-0.2cm}
  \caption{\label{fig:realworld_304} Real-world navigation experiment traversing through a glass door. (a) A scene photo captured during the experiment. (b) The unsafe trajectory generated by a standard perception and planning system. (c) The final executed trajectory and volumetric map using the proposed method, with glass surface indicated in light yellow and the touch action marked in pink.}
  \vspace{-0.8cm}
\end{figure}

We conduct a series of field experiments to validate the proposed framework for autonomous navigation in environments with transparent glass obstacles.
In the first scenario, a glass door frame is placed horizontally in a corridor between the start and goal positions of the aerial robot, as shown in Fig.~\ref{fig:realworld_304}(a). 
Using a standard perception and trajectory generation system, the aerial vehicle fails to perceive the glass surface, leaving a hole in the volumetric map. 
Consequently, it attempts to fly through the door frame along a minimum-time trajectory, as depicted in Fig.~\ref{fig:realworld_304}(b).
This unsafe trajectory would inevitably lead to a crash when colliding with the glass surface, which is not executed due to safety concerns.
In contrast, our proposed system with glass awareness actively engages with the detected glass surface by performing a touch action.
After confirming the presence of the glass surface, the aerial vehicle dynamically replans its trajectory, circumventing the obstacle by flying above the glass door frame, as shown in Fig.~\ref{fig:realworld_304}(c).
The entire trajectory including the touch action spans $6.58$ m and is completed in $24.23$ seconds.
This experiment demonstrated that when a glass surface obstructs the navigation path, the proposed system can effectively detect and engage contact with the glass surface for safe navigation in real-world scenarios.

\subsection{Navigating a Space with Two Glass Surfaces}
\label{subsec:exp_two_glass}

The second scene involves a room with dimensions $8 \times 7 \times 2$ m$^3$, which is separated by two fixed glass windows with half of the room inaccessible, as shown in Fig.~\ref{fig:realworld_combined}(1a).
The aerial vehicle is initially placed on one side of the room, with the goal position set on the opposite side \rev{(the red dot in Fig.~\ref{fig:realworld_combined}(1c))}.
During the experiment, the aerial vehicle attempts to reach the goal by flying through the right glass window, triggering the first touch action.
After confirming and updating the volumetric map with the right glass window, the aerial vehicle replans a new trajectory through the left glass window.
A second touch action is triggered there and the left window is also marked occupied.
At this point, no feasible trajectory to the goal position can be resolved, and the aerial vehicle terminates the task and hovers in a safe location.
The navigation task with a trajectory length of $9.48$ m is completed in $27.56$ seconds.
The data from the contact-sensing modules during the experiment are plotted in Fig.~\ref{fig:realworld_combined}(1b). 
This experiment highlights the system's ability to detect multiple glass surfaces and autonomously terminate the task when no feasible path is available, ensuring safety in complex environments.

\subsection{Navigating a Space with a Double Sliding Window}
\label{subsec:exp_window}

\begin{figure}[t] 
	\centering 
  \footnotesize
  \stackunder[5pt]{\includegraphics[width=0.48\columnwidth]{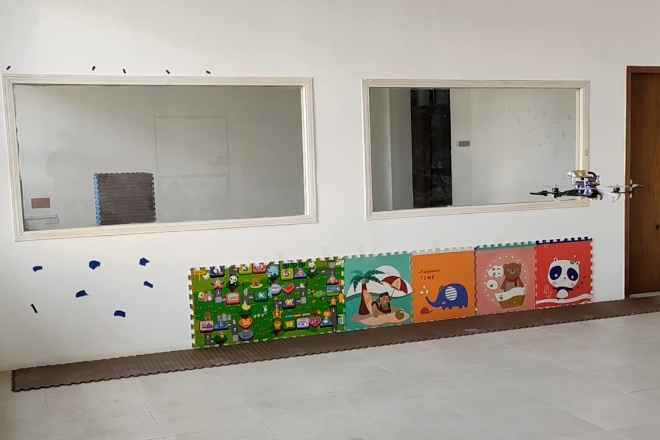}}{(1a)}%
  \hspace{0.15cm}%
  \stackunder[5pt]{\includegraphics[width=0.48\columnwidth]{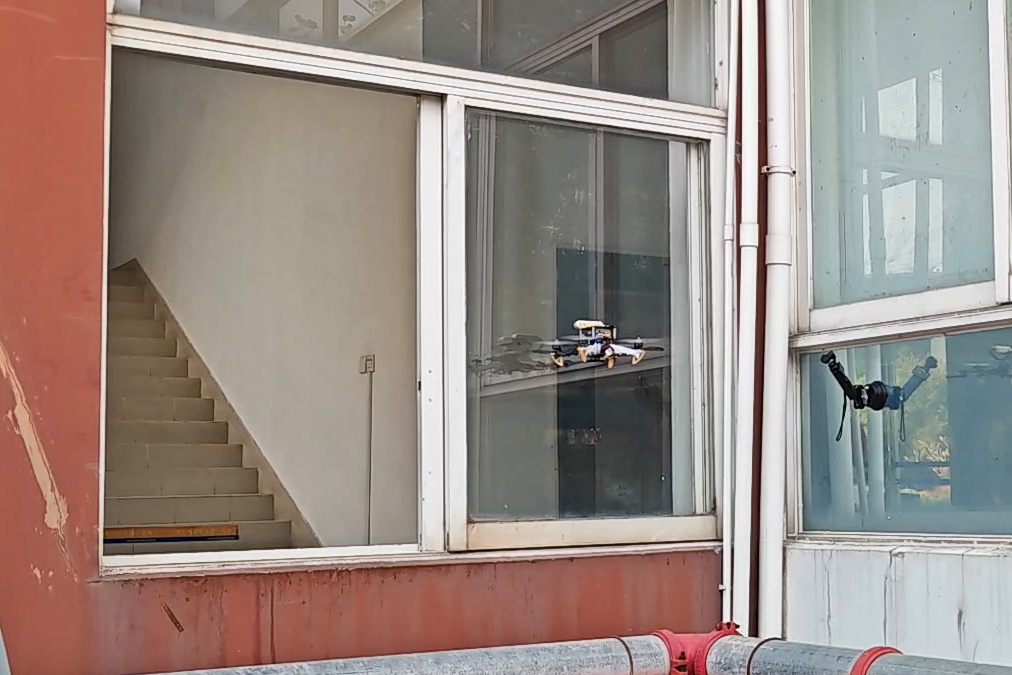}}{(2a)}%
  \vspace{0.15cm}
  \stackunder[5pt]{\includegraphics[width=0.48\columnwidth]{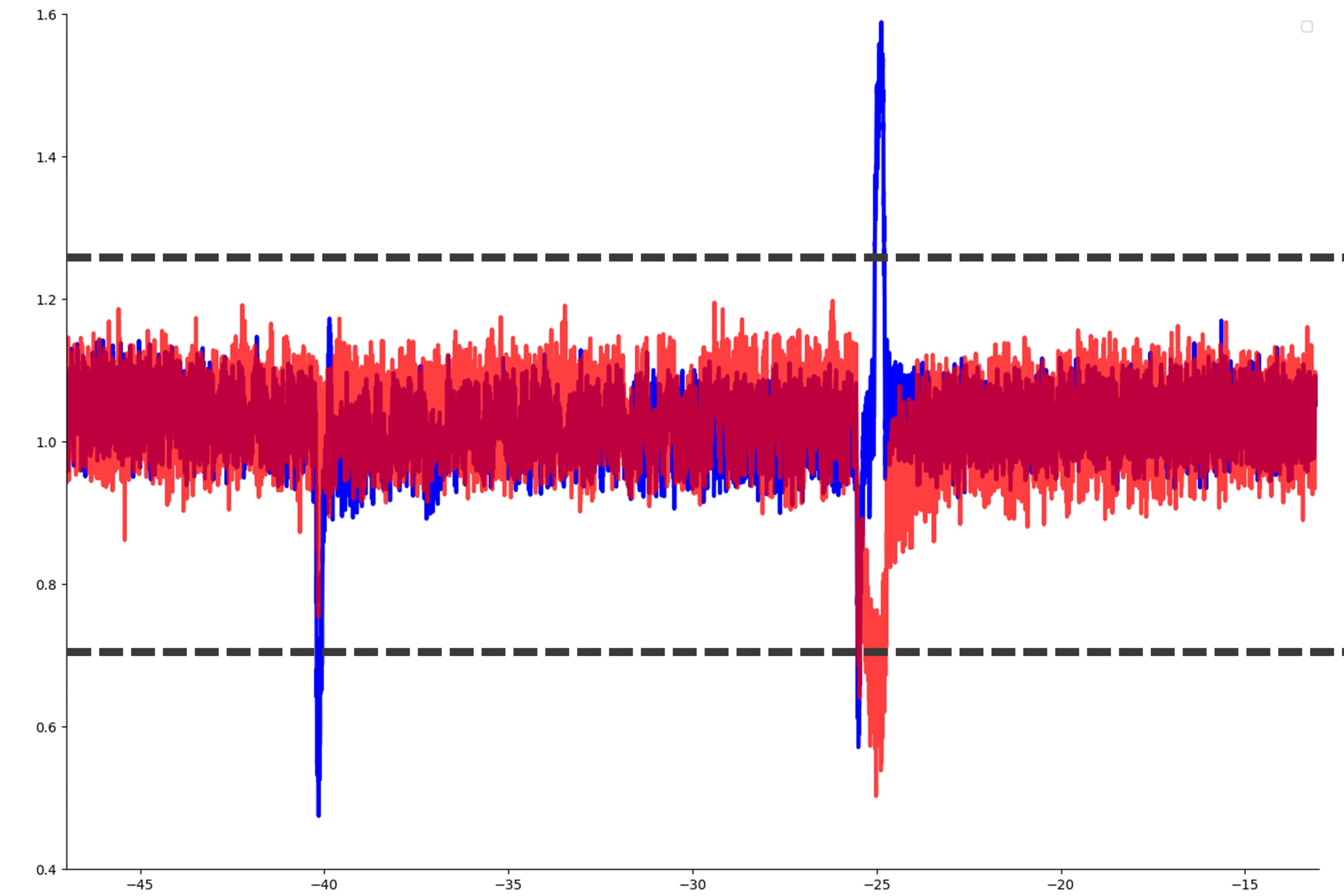}}{(1b)}%
  \hspace{0.15cm}%
  \stackunder[5pt]{\includegraphics[width=0.48\columnwidth]{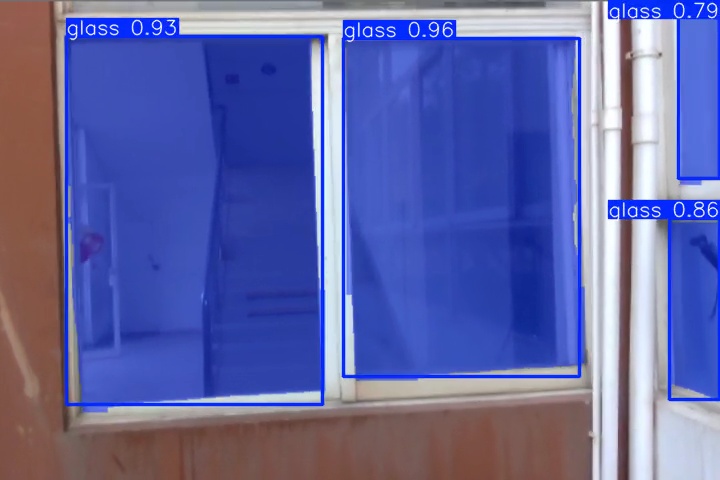}}{(2b)}%
  \vspace{0.15cm}
  \stackunder[5pt]{\includegraphics[width=0.48\columnwidth]{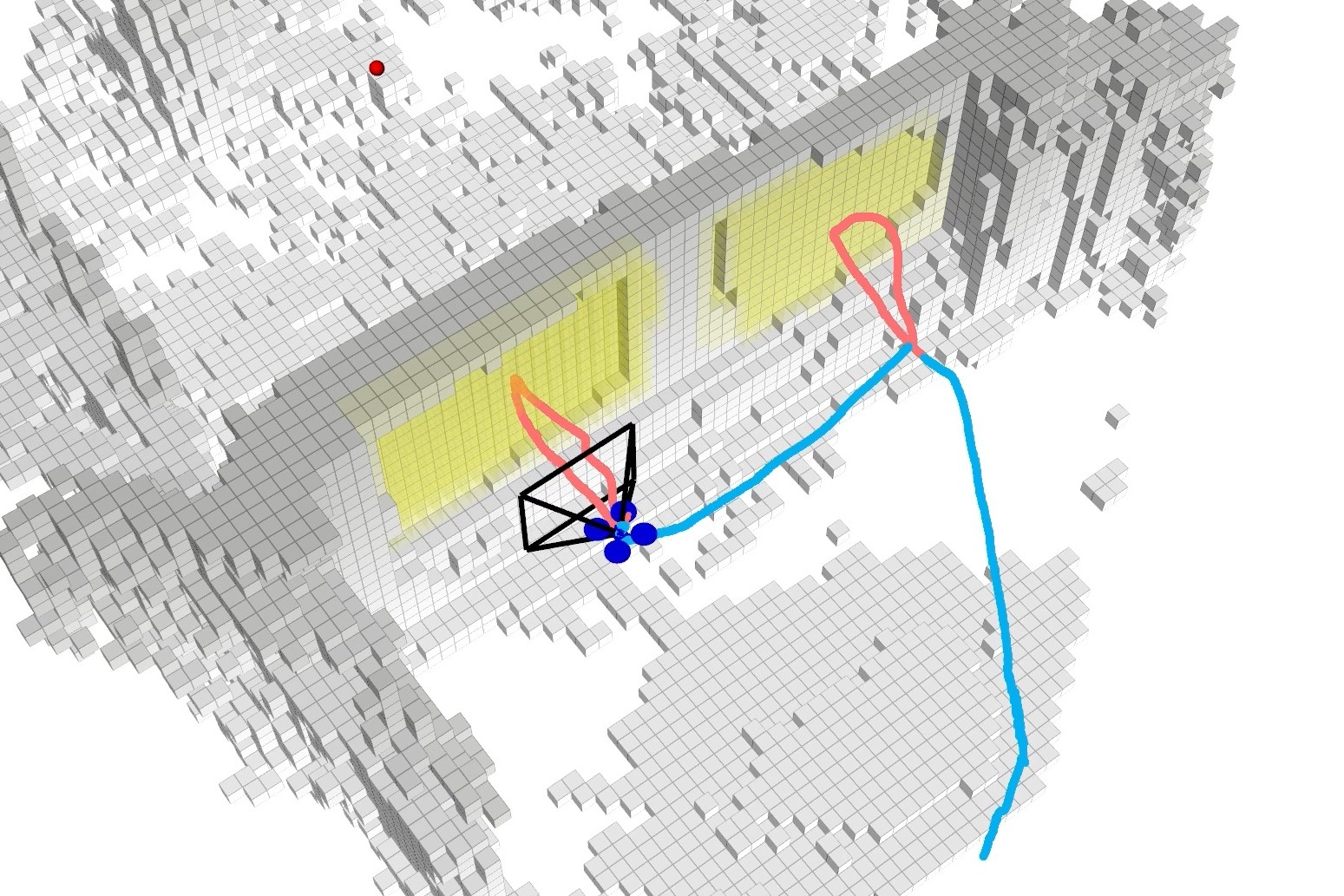}}{(1c)}%
  \hspace{0.15cm}%
  \stackunder[5pt]{\includegraphics[width=0.48\columnwidth]{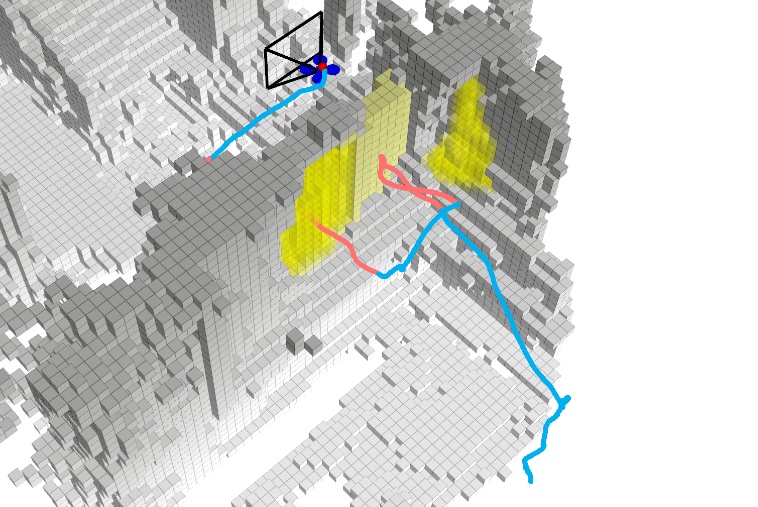}}{(2c)}%
  \caption{\label{fig:realworld_combined} Real-world navigation experiments (1) in a room with two fixed glass surfaces and (2) traversing through a double sliding window. (1a)(2a) Scene photos captured during the experiment. (1b) Data from the two contact-sensing modules, where the peaks exceeding the dashed black lines indicate touch action confirmations. (2b) A first-person view from the aerial vehicle, showing partially correct glass detection and segmentation. (1c)(2c) The final executed trajectory and volumetric mapping.}
  \vspace{-0.6cm}
\end{figure}

We conduct a more challenging point-to-point navigation experiment involving a double sliding glass window, as shown in Fig.~\ref{fig:realworld_combined}(2a).
The glass window is fully open, with the left half unobstructed and right half blocked by transparent glass.
As illustrated in Fig.~\ref{fig:realworld_combined}(2b), the onboard visual detection module cannot distinguish between the two halves of the window, identifying both regions as potential glass surfaces.
A standard trajectory is initially generated towards the goal position, which intersects with the right half of the window.
The aerial vehicle actively performs a touch action to confirm the presence of the glass surface and the volumetric map is updated by filling in the space occupied by the glass surface, as depicted in Fig.~\ref{fig:realworld_combined}(2c).
The planner then generates a new trajectory to the goal position using the updated volumetric map.
During this phase, an intersection is detected between the new trajectory and the left half of the window during the second safety check, prompting a second touch action.
The contact-sensing module does not register any contact this time, thereby invalidating the left glass surface.
Finally, the aerial vehicle replans one last trajectory and successfully reaches the goal position with a total trajectory length $8.86$ m and execution time $33.04$ s.
In this navigation task, the aerial vehicle successfully traverses the double sliding glass window by actively engaging with potential glass surfaces, even when the visual detection module provides only partially accurate information. 
If the agent relied solely on visual glass detection without the capability of physical contact, the task would fail due to the high confidence scores of both halves of the window (Fig.~\ref{fig:realworld_combined}(2b)).
The three navigation experiments in Sec.~\ref{subsec:exp_glass}~-~\ref{subsec:exp_window} demonstrate the effectiveness of the proposed system in achieving safe and robust navigation in complex and real-world scenarios with transparent glass obstacles.

\section{Conclusions}
\label{sec:conclude}
In this work, we propose a novel approach for robust autonomous aerial navigation in unknown environments with transparent glass obstacles.
Potential glass surfaces are detected and their information is incrementally maintained using visual sensor measurements.
The aerial vehicle then actively engages in touch actions with these surfaces using a pair of lightweight contact-sensing modules to confirm or invalidate their presence.
We integrate the proposed system into the fundamental task of autonomous point-to-point navigation in unknown environments.
We validate the incremental glass detection module in a glass-rich corridor and assess the advantage of active contact planning in simulation benchmarks.
A series of real-world experiments conducted in various scenarios demonstrate the system's reliability and practicality for safe and robust autonomous aerial navigation in complex environments with glass obstacles.

\addtolength{\textheight}{0.cm}   

\newlength{\bibitemsep}\setlength{\bibitemsep}{0.0\baselineskip}
\newlength{\bibparskip}\setlength{\bibparskip}{0.1pt}
\let\oldthebibliography\thebibliography
\renewcommand\thebibliography[1]{%
\oldthebibliography{#1}%
\setlength{\parskip}{\bibitemsep}%
\setlength{\itemsep}{\bibparskip}%
}

\bibliography{ref}

\end{document}